\documentclass[letterpaper, 10 pt, conference]{ieeeconf}  

\IEEEoverridecommandlockouts                              

\usepackage{times}
\usepackage{epsfig}
\usepackage{graphicx}
\usepackage{amsmath}
\usepackage{amssymb}

\usepackage{multirow}
\usepackage{xcolor}
\usepackage{footnote}
\usepackage{microtype}

\newcommand{\lowerbetter}{\color{red} \mathbf{\downarrow}}
\newcommand{\higherbetter}{\color{red} \mathbf{\uparrow}}

\newcommand{\rmse}{\text{RMSE} \lowerbetter}
\newcommand{\rmselog}{\text{RMSE}_{\log} \lowerbetter}

\newcommand{\absrel}{\text{Abs Rel} \lowerbetter}
\newcommand{\sqrel}{\text{Sq Rel} \lowerbetter}

\newcommand{\lidar}{LiDAR}

\newcommand{\radar}{Radar}

\definecolor{Highlight}{HTML}{39b54a}
\newcommand{\improves}[1]{{ \color{Highlight} {(#1)}}}

\title{Depth Estimation Matters Most: Improving Per-Object Depth Estimation \\ for Monocular 3D Detection and Tracking}

\author{Longlong Jing$^{1}$, Ruichi Yu$^{1}$, Henrik Kretzschmar$^{1}$, Kang Li$^{1}$, Charles R. Qi$^{1}$, Hang Zhao$^{1*}$, Alper Ayvaci$^{1}$ \\ Xu Chen$^{1}$, Dillon Cower$^{1}$, Yingwei Li$^{2}$, Yurong You$^{3}$, Han Deng$^{1}$, Congcong Li$^{1}$, and Dragomir Anguelov$^{1}$\\
$^{1}$Waymo LLC, $^{2}$Johns Hopkins University, $^{3}$Cornell University
\thanks{$^{*}$Work done while at Waymo LLC.}%
}

\begin{document}

\maketitle
\thispagestyle{empty}
\pagestyle{empty}

\begin{abstract}

Monocular image-based 3D perception has become an active research area in recent years owing to its applications in autonomous driving. Approaches to monocular 3D perception including detection and tracking, however, often yield inferior performance when compared to LiDAR-based techniques. Through systematic analysis, we identified that per-object depth estimation accuracy is a major factor bounding the performance. Motivated by this observation, we propose a multi-level fusion method that combines different representations (RGB and pseudo-LiDAR) and temporal information across multiple frames for objects (tracklets) to enhance per-object depth estimation. Our proposed fusion method achieves the state-of-the-art performance of per-object depth estimation on the Waymo Open Dataset, the KITTI detection dataset, and the KITTI MOT dataset. We further demonstrate that by simply replacing estimated depth with fusion-enhanced depth, we can achieve significant improvements in monocular 3D perception tasks, including detection and tracking.

\end{abstract}

\section{Introduction}

Existing perception systems for autonomous vehicles mainly rely on expensive sensors such as \lidar{} and \radar{} \cite{lang2019pointpillars, liu2019flownet3d, behley2019semantickitti, yang2020radarnet}. Owing to the low cost, low power consumption and longer perception range of cameras, monocular image-based perception has been attracting great interest in recent years from both the industry and the research community \cite{chen2020monopair, alhashim2018high, wang2019pseudo, ma2020rethinking, zhou2019objects, weng20203d, hu2019joint}. Such perception tasks tend to be challenging, and there is a large performance gap between monocular perception systems and LiDAR/radar-based systems~\cite{wang2019pseudo, zhou2019objects, chen2020monopair, wang2020centernet3d}.

\begin{figure}
\begin{center}
\includegraphics[width=0.89\linewidth]{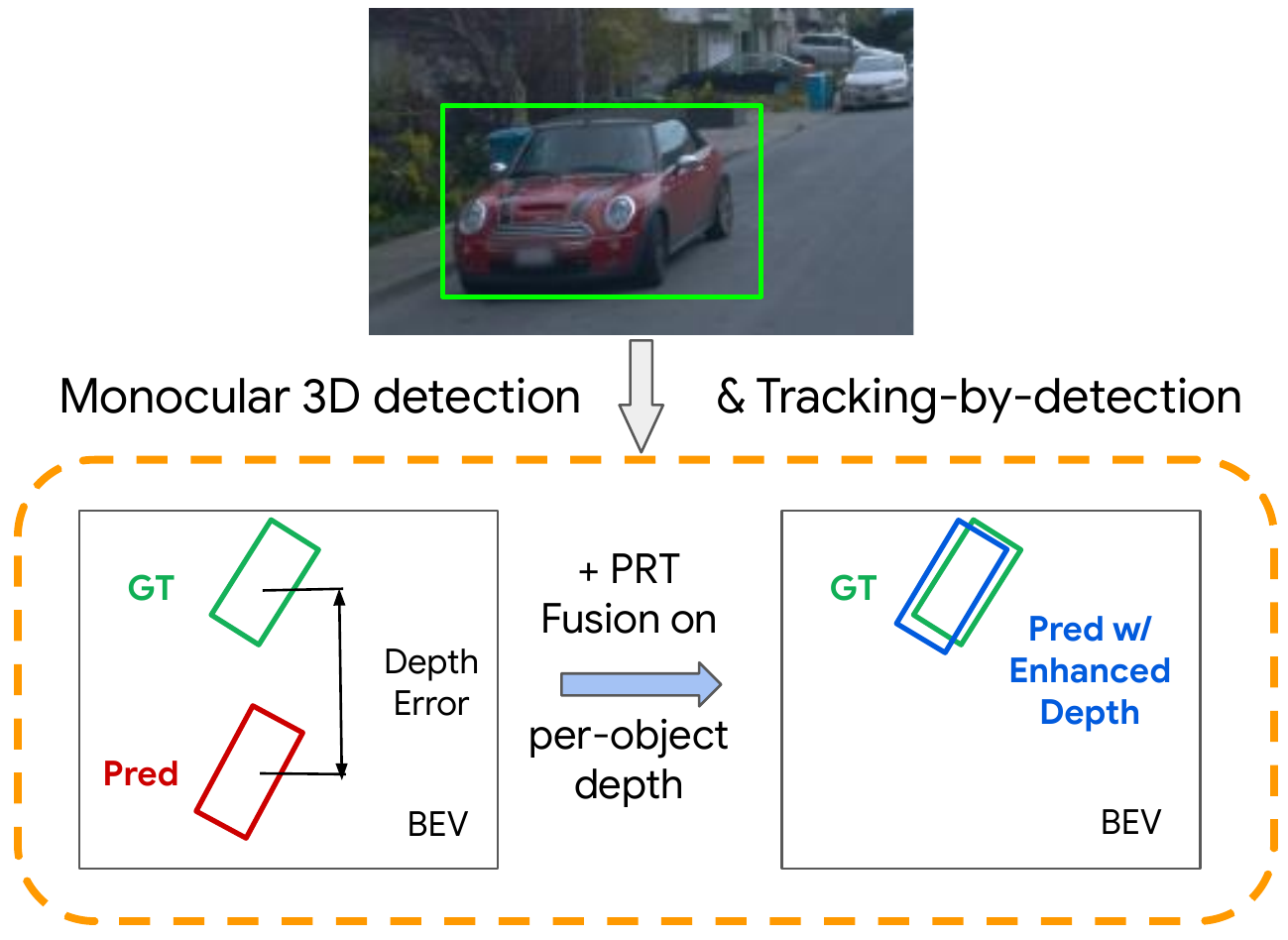}
\end{center}
   \caption{Illustration of the impact of object depth estimation for monocular 3D detection. Predicted box (Pred) and ground truth (GT) box are visualized in a bird's-eye-view (BEV). In the bottom left image, we see depth discrepancy causes localization error. In the bottom right image, we see our proposed \underline{P}seudo-LiDAR, \underline{R}GB and \underline{T}racklet (PRT) fusion method can improve object depth estimation and improve detection.}
\label{fig:overall}
\end{figure}

\begin{figure}
\begin{center}
\includegraphics[width=0.9\linewidth,height=5cm]{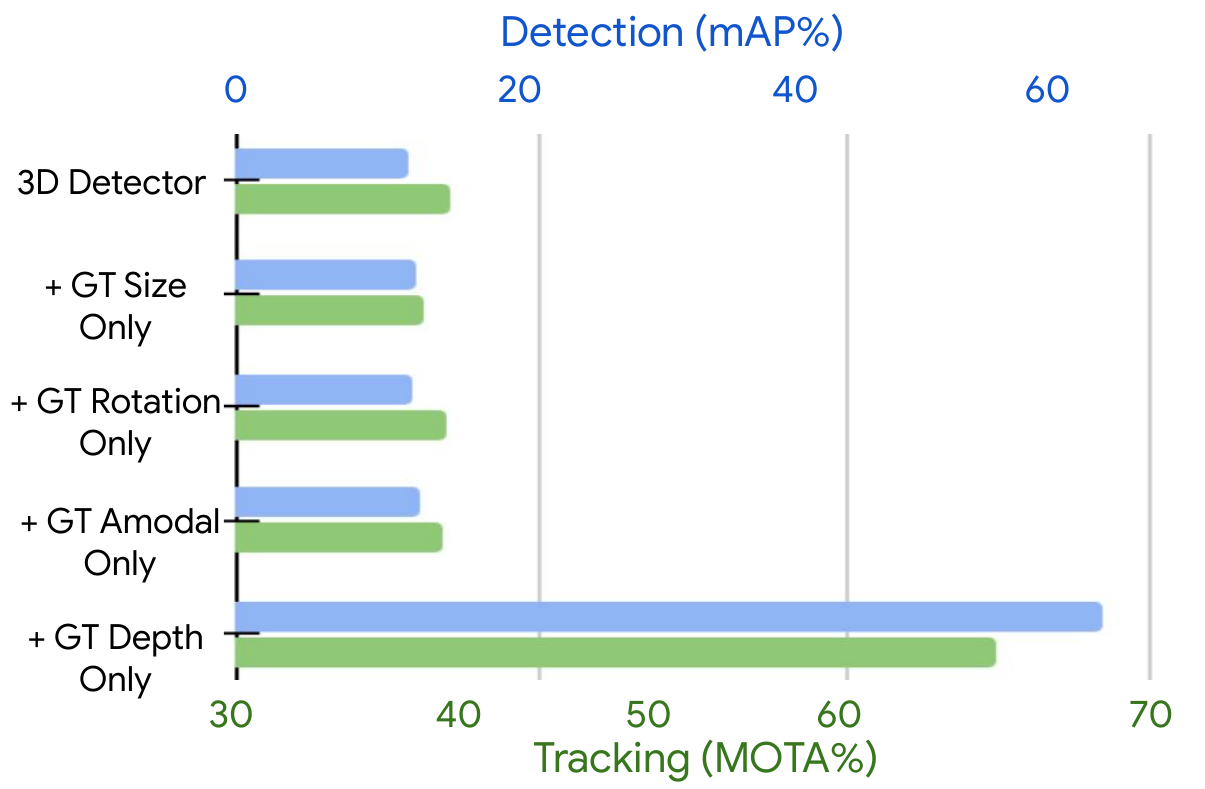}
\end{center}
   \caption{Headroom analysis of the impact of each component of the monocular 3D detection and tracking system. The experiment is done with the state-of-the-art monocular 3D detector CenterNet \cite{zhou2020tracking} (ranked 1st on nuScene dataset~\cite{caesar2020nuscenes}) and the AB3D~\cite{weng20203d} tracker. We use ``+ GT" to indicate that we replaced the prediction with the ground truth. The analysis suggests that depth has the most significant impact on detection (Average Precision shown as the top bar) and tracking (MOTA shown as the bottom bar) performance.}
\vspace{-10pt}
\label{fig:motivation}
\end{figure}

Common 3D monocular perception systems comprise two major modules: 3D object detection and 3D tracking\footnote{In this paper we follow the tracking-by-detection paradigm.}. The former requires learning the 3D location, box size, and rotation/orientation of an object, while the latter requires using appearance and motion cues to track detections across frames. Among both tasks, it is not obvious which component of the system has the most crucial impact on performance. To fully understand which component bounds the overall performance, we experimented with replacing each output from a state-of-the-art detection model with the ground truth and then evaluated changes of the detection
and tracking-by-detection performance using the state-of-the-art detector. As shown in Fig.~\ref{fig:motivation}, among all the attributes including rotation, size, depth, and amodal box center in image, we find that only the per-object depth, the depth of the vehicle's 3D center, matters (see the significant performance improvements when per-object depth is perfect, and marginal improvements when other signals are perfect). Based on this observation, we identified that per-object depth estimation is a major bottleneck for monocular 3D detection and tracking-by-detection. We also conducted the same analysis with other state-of-the-art detectors such as RTM3D~\cite{RTM3D} with the AB3D~\cite{weng20203d} tracker, and the results suggest that depth is the key factor to improve monocular 3D detection and tracking is a general conclusion across models. In this paper, we focus on improving per-object depth estimation and demonstrate that by just enhancing object depth we can significantly improve detection and tracking performance.

A major challenge of estimating object depth from a monocular image is to obtain a representation that encodes the transition from 2D information to 3D depth. Recent efforts (e.g., 3D monocular detection) have mainly focused on either directly learning from the raw RGB image \cite{zhou2019objects, chen2020monopair, liu2020smoke} or leveraging a pseudo-LiDAR representation lifted from the predicted dense depth map \cite{wang2019pseudo, ma2020rethinking, vianney2019refinedmpl}. 
Intuitively, we believe that the above two representations might be complementary in estimating per-object depth, and learning from either one of them alone might be sub-optimal: the RGB image encodes the appearance, texture and 2D geometry, etc. of an object explicitly but contains no direct information of 3D. It is difficult to learn how to map RGB features to depth precisely without overfitting to irrelevant information; on the other hand, the pseudo-LiDAR representation directly models the 3D structure of an object via an estimated dense depth map, which makes it straightforward to learn per-object depth. However the estimated dense depth map is often noisy (usually with at least $8$\% average relative error \cite{alhashim2018high, godard2017unsupervised, godard2019digging}). Inspired by previous methods that fuse different representations such as RGB~image features and optical flow for action recognition~\cite{simonyan2014two}, we believe that fusing the complementary signals encoded in the two representations may help per-object depth estimation.

Furthermore, depth estimation from monocular images is fundamentally ill-posed, as a single 2D view of a scene can be explained by many plausible 3D scenes~\cite{qi2018frustum}. However, observing an object over time allows us to model the underlying temporal and motion consistency of the object, which can provide contextual information to better localize the object in 3D{\color{red}$^{\:2}$}. Similar ideas have been explored in other tasks such as 2D video-based object detection \cite{beery2020context, liu2019looking, zhu2017flow}. 

Based on our intuitions and the analysis above, we propose a multi-level fusion framework that consists of two major components. The first component is the pseudo-LiDAR and RGB fusion (PR-Fusion) which enhances depth estimation from two complementary representations. The other component is the tracklet fusion (T-fusion), which leverages temporal and motion consistency with compensated ego-motion. Our full model \underline{P}seudo-LiDAR-\underline{R}GB-\underline{T}racklet (PRT fusion) fuses information from both 2D and 3D representations across multiple frames.

We conducted extensive experiments on the Waymo Open Dataset~\cite{sun2020scalability}, the KITTI detection dataset~\cite{geiger2013vision}, and the KITTI MOT dataset~\cite{geiger2013vision} to demonstrate the effectiveness of our method. We obtained state-of-the-art performance on the per-object depth estimation benchmark proposed by \cite{zhu2019learning}. To further demonstrate the practical value of enhanced per-object depth, we have also achieved significant improvements on 3D object detection and tracking models, by simply replacing the per-object depth with our enhanced estimation.

Our contributions can be summarized as follows: 1) We conducted a systematic analysis identifying that per-object depth estimation is a major performance bottleneck of current 3D monocular detection and tracking-by-detection methods.
2) We proposed a novel method that fuses pseudo-LiDAR and RGB information across the temporal domain to significantly enhance per-object depth estimation performance.
3) We demonstrated that with the enhanced depth, the performance of monocular 3D detection and tracking can be significantly improved.

\section{Related Work}

\textbf{Monocular 3D Object Detection and Tracking:} Monocular 3D object detection models aim at directly regressing attributes such as rotation, size, and depth from RGB images \cite{chen2020monopair, zhou2019objects, zhou2020tracking, liu2020smoke, brazil2019m3d, simonelli2019disentangling, jorgensen2019monocular} or pseudo-LiDAR~\cite{wang2019pseudo,vianney2019refinedmpl, weng2019monocular, ma2020rethinking}. Recently, Wang \textit{et al.} pointed out that image features are not suitable for the task; instead, they proposed converting image-based depth maps into pseudo-LiDAR representations to mimic LiDAR and obtained significantly better performance than the previous image-based methods~\cite{wang2019pseudo}. A few work attempted to leverage both the RGB image and estimated depth map for monocular 3D object detection \cite{ding2020learning, ma2019accurate}; none of them specifically focused on improving the per-object depth estimation based on different features from a tracklet. Most existing tracking methods follow the tracking-by-detection scheme \cite{zhou2020tracking, weng20203d, bewley2016simple, weng2020gnn3dmot}, and the quality of monocular 3D detections is the bottleneck for tracking performance. Among all of the output dimensions of the above tasks, we identified that per-object depth is the bottleneck and demonstrated that the performance of monocular 3D detection and tracking can be significantly improved with enhanced object depth.

\textbf{Depth/Distance Estimation:} Monocular image-based dense depth estimation has been studied for years \cite{alhashim2018high, eigen2014depth, godard2019digging, godard2017unsupervised, guizilini2020semantically, wang2020task, poggi2020uncertainty, fu2018deep}. Different from the existing methods, our method focused on per-object depth estimation, which naturally enables the novel tracklet-based fusion which fuses different types of features across multiple frames. Although the two tasks might share some high-level concepts (depth estimation), it is non-trivial to adapt the per-pixel dense depth estimation to a per-object task, which has not been explored in the dense depth estimation community. On the other hand, per-object depth/distance estimation has started to draw attention recently. Following the same experimental setting in of the most recent state-of-the-art~\cite{zhu2019learning}, our novel framework, which leverages both pseudo-LiDAR and RGB representations across multiple frames for per-object depth estimation, demonstrated superior performance.

\begin{figure*}
\begin{center}
\includegraphics[width=0.9\linewidth]{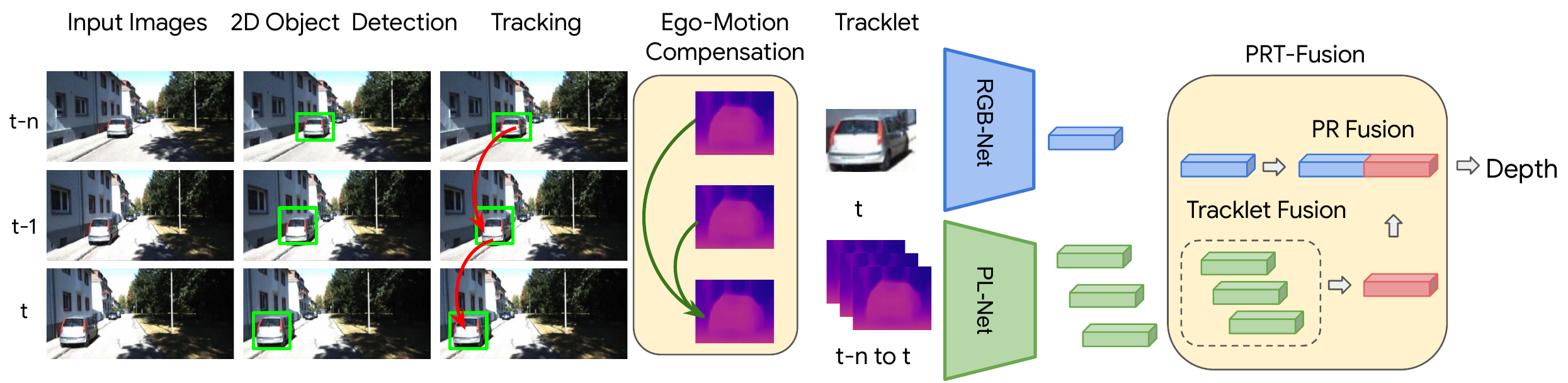}
\end{center}
   \caption{The overall framework for our proposed monocular per-object depth estimation method. The entire framework includes three stages: (1) per-frame 2D object detection to detect objects, (2) 2D tracking to associate objects for the same vehicle across the temporal domain, (3) the proposed PRT-Fusion over the tracklets generated by the tracking methods, with fusion from both the temporal domain and different representations (RGB and pseudo-LiDAR).
   }
\label{fig:framework}
\vspace{-10pt}
\end{figure*}

\textbf{Representation and Temporal Fusion} Many existing methods are based on representation \cite{vora2020pointpainting, liang2019multi, liang2018deep, fadadu2020multi, xiao2020audiovisual, hazirbas2016fusenet} and temporal fusion \cite{chen2020monopair, alhashim2018high, wang2019pseudo, ma2020rethinking, zhou2019objects, weng20203d, hu2019joint}. Two-stream fusion \cite{simonyan2014two, feichtenhofer2016convolutional} for action recognition is a classic multi-modal fusion which fuses RGB image features and optical flow \cite{dosovitskiy2015flownet}. In the temporal domain, many methods have been explored to improve sequence-based tasks such as video object detection and video segmentation \cite{zhou2019objects, beery2020context, liu2019looking, zhu2017flow}. These methods inspired us that different representations (e.g., pseudo-LiDAR and RGB) across multiple frames might be beneficial for improving per-object depth estimation.

\section{Improving Per-object Depth Estimation via Multi-level Fusion}

The overview of our proposed multi-level fusion framework for per-object depth estimation is shown in Fig.~\ref{fig:framework}. We first conduct 2D object detection and track detections across frames to construct a tracklet for each object. We then construct pseudo-LiDAR representations of the objects across frames and RGB image features for the current frame. Ego-motion compensation is applied to all pseudo-LiDAR patches within each tracklet to transform them to the same coordinate system. Finally, the RGB image features for the current frame and the temporally fused pseudo-LiDAR features are fused to produce per-object depth.

\subsection{Pseudo-LiDAR and RGB (PR) Fusion}
\label{sec:per-frame-fusion}

Inspired by the two-stream fusion method for action recognition presented in \cite{simonyan2014two}, we proposed PR-Fusion to leverage the complementary information encoded from both RGB and pseudo-LiDAR representations. Given an RGB image $\textbf{I}$ with size $H \times W$, compact features for the entire image can be extracted by using a pre-trained convolution neural network $F_{RGB}$. For any object with its 2D bounding box $\textbf{b}$, the RGB image features $\textbf{R}$ for the bounding box can be extracted by using the pre-defined pooling operation $\text{Pool}(F_{\text{RGB}}(\textbf{I}), \textbf{b})$. The process of extracting image features $\textbf{R}$ for object bounding box $b$ from image $I$ in can be represented as 
\begin{equation}
\textbf{R} = \text{Pool}(F_{\text{RGB}}(\textbf{I}), \textbf{b}).
\end{equation} 

The extraction process of the pseudo-LiDAR representation consists of three steps: (1) dense depth estimation for each image, (2) lifting predicted dense depth into pseudo-LiDAR, and (3) pseudo-LiDAR representation extraction with a neural network. For any RGB image $I$, the depth estimation can be accomplished by using a dense depth estimation network $F_d$ as
\begin{equation}
\textbf{d} = F_d(\textbf{I}),
\end{equation}while \textbf{d} is a 2D dense depth estimation map having the same size as input image $I$. The pixel value at location $(u, v)$ of the $\textbf{d}$ indicates the depth of the corresponding pixel in the image.

Then each pixel of the entire depth map is lifted into a point cloud by using the following equations based on the camera model:
\begin{equation}
\left\{
      \begin{array}{lr}
      z = \textbf{d}(u, v) ,\\
      x = (u-C_{x}) \times z / f_{x}  ,\\
      y = (v-C_{y}) \times z / f_{y}  ,\\
      \end{array}
\right.
\label{eq:transformation}
\end{equation}
where $(f_{x}, f_{y})$ is the horizontal and vertical focal lengths of the camera and $(C_{x}, C_{y})$ is the pixel location corresponding to the camera center \cite{ma2020rethinking, wang2019pseudo}. After the transformation, each pixel in the dense depth map $\textbf{d}$ is transformed into three channels representing the absolute location of the corresponding pixel in 3D space, in camera coordinates. 

After obtaining the pseudo-LiDAR representation for image $\textbf{I}$, the pseudo-LiDAR patch $\textbf{P}_t$ for object $\textbf{b}_t$ at timestamp $t$ can be cropped based on the 2D bounding box, where $\textbf{P}_t$ is a collection of pseudo-LiDAR points that are within the box $\textbf{b}_t$. The pseudo-LiDAR-based features $\textbf{PL}$ of object $\textbf{b}_t$ can be extracted using another feature encoder $F_p$ as

\begin{equation}
\textbf{PL}  = F_p(\textbf{P}_t),
\end{equation}where $\textbf{PL}$ represents the pseudo-LiDAR representation for the object within the bounding box $b_t$ in the image plane. 

Finally, the PR-Fusion can be represented as
\begin{equation}
\label{eq:per-frame_fusion}
\textbf{PR}  = G_{\text{PR}}(\textbf{PL} , \textbf{R} ), 
\end{equation}where $G_{\text{PR}}$ is a deep neural network to fuse the two features and $\textbf{PR}$ is the fused feature.

\subsection{Tracklet Fusion with Ego-motion Compensation}

\label{sec:temporal-fusion}

Predicting per-object depth directly from a single frame is challenging due to the fact that a single object in an camera image can be explained by multiple plausible objects with different depth \cite{qi2018frustum}. Inspired by temporal fusion methods for video based tasks, we propose to fuse the object-level information across multiple frames to enforce temporal and motion consistency of the prediction. Given the 2D detection results, we first conduct 2D data association \cite{bewley2016simple, weng20203d} to construct tracklets for objects and then fuse the features of the tracklet in a temporal window.

A straightforward method is to directly fuse image features across frames similar to \cite{zhu2017flow}; however, we find that directly fusing the RGB features from different frames can be suboptimal because RGB features couple the dynamic motion of the camera and the motion of the objects together, which makes it hard to learn motion and temporal consistency from 2D image sequence. We believe that to perform effective temporal fusion for depth estimation, the camera motion must be compensated to make sure the features from different frames are in the same coordinate system. Fortunately, the ego-motion of the camera can be easily compensated in the 3D space with the pseudo-LiDAR representations. Thus, we propose a T-Fusion method with ego-motion compensation based on pseudo-LiDAR representations.

The input to our proposed T-Fusion includes pseudo-LiDAR patches of each object in different frames $\textbf{P}_t, \textbf{P}_{t-1}, ..., \textbf{P}_{t-n}$, while $\textbf{P}_t$ is in the 3D camera coordinate in frame $t$. The ego-motion is represented using a $4 \times 4$ homogeneous matrix $\textbf{H}$ based on conventional six degrees of freedom: translation $[\gamma_x, \gamma_y, \gamma_z]$ in meters and rotation $[\rho_x, \rho_y, \rho_z]$ in radians. 

{
First, all pseudo-LiDAR patches from different frames are projected into the global coordinate system using the camera-coordinate-to-global-coordinate transformation matrix $\textbf{H}$. For a pseudo-LiDAR patch for any timestamp $\textbf{P}_{t-j}$, assume its camera-coordinate-to-global-coordinate transformation matrix is $\textbf{H}_{t-j}$; the transformation is as follows:
\begin{equation}
    \textbf{P}_{t-j}' = \textbf{H}_{t}^{-1} * \textbf{H}_{t-j} * \textbf{P}_{t-j}.
\end{equation}
After the coordinate transformation, the ego-motion of the self-driving car is compensated for, and the transformed $P_{t-j}'$ is in the same coordinate system as $P_t$. The same transformation is applied to pseudo-LiDAR patches from all timestamps to eliminate the impact of ego-motion to the locations of pseudo-LiDAR points for each object.

Given any feature encoder $F_p()$ for pseudo-LiDAR, the features for different timestamps of data can be extracted as $F_p(\textbf{P}'_t)$, $F_p(\textbf{P}'_{t-1})$, ..., $F_p( \textbf{P}'_{t-n})$, where the $'$ indicates that the pseudo-LiDAR patch has ego-motion compensated. Then, the fused features for a sequence of an object can be modeled by using a neural network encoder $G_{\text{TF}}$ as follows: 
\begin{equation}
\textbf{PL}_{t-n \rightarrow t} = G_{\text{TF}}(F_p(\textbf{P}'_{t}), F_p(\textbf{P}'_{t-1}), ...., F_p(\textbf{P}'_{t-n})),
\end{equation} where $\textbf{PL}_{t-n \rightarrow t}$ is the fused tracklet features from frame $t-n$ to $t$.
}

\subsection{Multi-level PRT-Fusion}
\label{sec:multi-level-fusion}

{
PR-Fusion and T-Fusion aggregate features from two different domains. It is natural to combine the two fusion methods together for further performance improvements. Given a sequence of object boxes across time, $\textbf{b}_{t}, \textbf{b}_{t-1}, ...., \textbf{b}_{t-n}$, the RGB image features for object $b_i$ can be represented using an image feature encoder $F_{\text{RGB}}()$, and its pseudo-LiDAR features can be extracted using encoder $F_p()$. There are two steps in PRT-Fusion: Given the object in the current frame and its previous frames, first we conduct T-Fusion with ego-motion compensation for pseudo-LiDAR representations across multiple frames; then we fuse it with the RGB features at the current frame $t$ as
\begin{equation} 
\textbf{PRT} = G_{\text{PR}}(\textbf{PL}_{t-n \rightarrow t}, \textbf{R}_t),
\end{equation}where \textbf{PRT} is the fused features from frame $t-n$ to $t$.
}

\subsection{Implementation Details}
{ 

\textbf{RGB Feature Extraction.} CenterNet and CenterTrack \cite{zhou2019objects, zhou2020tracking} have achieved state-of-the-art performance on monocular 3D detection task on the nuScenes dataset recently \cite{caesar2020nuscenes}. We followed its formulation and network architecture with ResNet50 \cite{he2016deep} as the backbone to perform 2D detections. 

\textbf{Pseudo-LiDAR Feature Extraction.} Recently, PatchNet \cite{ma2020rethinking} was proposed to significantly improve pseudo-LiDAR-based detection performance. We choose it as our backbone model to extract pseudo-LiDAR-based features as both the baseline and the input to our method.

\textbf{2D Tracking.} To track 2D detections to form tracklets, we followed \cite{bewley2016simple} to use a Kalman-Filter based tracker. It is worth noting that since our paper mainly focused on per-object depth estimation, we believe that with more sophisticated tracking methods \cite{zhang2020fair, hung2020soda, wang2019towards}, the performance of our fusion method can be further improved.

}

\section{Experiments}

In this section, we firstly benchmark our per-object depth estimation by comparing with prior works in Sec.~\ref{sec:exp:benchmark}. Next, we ablate design choices in our depth estimation model in Sec.~\ref{sec:exp:ablation}. Finally, we show the applications of the improved per-object depth for 3D monocular detection and tracking in Sec.~\ref{sec:exp:app}.

\textbf{Datasets.} We evaluate on multiple datasets with a focus on the vehicle class. Among them, \emph{Waymo Open Dataset} is a large-scale dataset for autonomous driving. It consists of $798$ training sequences and $202$ validation sequences, while each sequence contains around $200$ frames. \emph{KITTI Detection Dataset} has $3,712$ RGB images for training and $3,768$ images for testing. We are using the split used in the prior work \cite{zhu2019learning} for per-object depth estimation for fair comparison. \emph{KITTI MOT Dataset} consists of $8,008$ and $11,095$ frames in the official training and testing splits. Since it has sequence information for each frame, we will use this dataset to demonstrate the effectiveness of the PRT fusion.

\subsection{Benchmarking Per-object Depth Estimation} 
\label{sec:exp:benchmark}

\textbf{Metrics}. Following the existing state-of-the-art per-object depth estimation benchmark proposed by \cite{zhu2019learning}, five standard metrics including average relative error (Abs Rel), squared relative error (Sq Rel), root-mean-square error (RMSE), average ($\log_{10}$) error ($\text{RMSE}_{\log}$), and threshold accuracy ($\delta_i$) are used for evaluation.

\begin{savenotes}
\begin{table}[t]
\centering
\caption{Comparison with state-of-the-art methods on the Waymo Open Dataset for vehicles following the setting in \cite{zhu2019learning}.}
\scalebox{0.9}{
\begin{tabular}{c|c|cccc}
\hline
Method & $\delta < 1.25 \higherbetter$  & $\absrel$ & $\sqrel$ & $\rmse$ & $\rmselog$ \\ 
\hline
SVR~\cite{gokcce2015vision}    &83.26\%   &14.79\% &1.3254 &6.9081 &0.2282\\
DistNet~\cite{haseeb2018disnet} &88.50\%  &11.23\% &0.8974 &6.3903 &0.1737\\
PatchNet~\cite{ma2020rethinking}    &92.81\%  &8.77\% &0.6051 &5.5485 &0.1283\\
CenterNet~\cite{zhou2019objects} &95.47\%  &7.25\% &0.6240 &4.7506 &0.1146\\
Ours (T) &96.45\%   &6.96\% &0.4214 &4.5941 &0.1001\\
Ours (PR) &97.64\%   &5.74\% &0.3188 &3.8788 &0.0863\\
Ours (PRT) &\textbf{98.09\%} &\textbf{5.47\%} &\textbf{0.2858} &\textbf{3.7282} &\textbf{0.0802}\\
\hline
\end{tabular}
}
\label{tab:compare-waymo}
\end{table}
\end{savenotes}

\begin{table}[t]
\centering
\caption{Performance comparison with state-of-the-arts on the KITTI Detection Dataset for vehicles following the setting in \cite{zhu2019learning}.}
\scalebox{0.9}{
\begin{tabular}{c|c|cccc}
\hline
Method & $\delta < 1.25 \higherbetter$  & $\absrel$ & $\sqrel$ & $\rmse$ & $\rmselog$ \\   
\hline
SVR  \cite{gokcce2015vision}   &34.50\%   &149.4\% &47.748 &18.970 &1.4940\\
IPM \cite{tuohy2010distance}    &70.10\%  &49.70\% &1290.5 &237.62 &0.4510\\
Zhu \textit{et al.} \cite{zhu2019learning}   &84.60\%  &15.00\% &0.6180 &3.9460 &0.2040\\
DistNet~\cite{haseeb2018disnet} &93.26\%  &12.39\% &0.4834 &2.9539 &0.2003\\
CenterNet~\cite{zhou2019objects} &95.33\%  &8.70\% &0.4250 &3.2433 &0.1436\\
PatchNet~\cite{ma2020rethinking} &95.52\%  &8.08\% &0.2789 &2.9048 &0.1296\\
Ours (PR) &\textbf{97.60\%} &\textbf{6.89\%} &\textbf{0.2340} &\textbf{2.5025} &\textbf{0.1181}\\
\hline
\end{tabular}
}
\label{tab:compare-kitti-detection}
\end{table}

\begin{table}[t]
\centering
\caption{Performance comparison with state-of-the-arts on the KITTI MOT Dataset for vehicles following the setting in \cite{zhu2019learning}.}
\scalebox{0.9}{
\begin{tabular}{c|c|cccc}
\hline
Method & $\delta < 1.25 \higherbetter$ & $\absrel$ & $\sqrel$ & $\rmse$ & $\rmselog$ \\ 
\hline
CenterNet~\cite{zhou2019objects} &92.17\%  &9.10\% &0.9372 &6.9596 &0.1975\\
PatchNet~\cite{ma2020rethinking} &93.41\%  &8.65\% &0.4988 &4.9081 &0.1268\\
Ours (T) &93.43\%  &7.90\% &0.3706 &4.0703 &0.1157\\
Ours (PR) &94.39\%   &7.69\% &0.4430 &4.8065 &0.1205\\
Ours (PRT) &\textbf{95.23\%}  &\textbf{7.13\%} &\textbf{0.3382} &\textbf{3.9391} &\textbf{0.1076}\\
\hline
\end{tabular}
}
\label{tab:compare-kitti-mot}
\end{table}

\textbf{Results}. Since our paper mainly focused on per-object depth estimation, we compare the performance of our proposed fusion methods against the state-of-the-art models. We compare with two types of methods: geometry based methods which predict the depth based on the geometry of the boxes including SVR~\cite{gokcce2015vision}, IPM \cite{tuohy2010distance}, and DistNet~\cite{haseeb2018disnet}, and deep feature based methods including the methods proposed in \cite{zhu2019learning}, and our implementation of state-of-the-art monocular 3D detection method CenterNet \cite{zhou2019objects} and PatchNet \cite{ma2020rethinking} (we simply used their backbone and change the final prediction from a 3D box to only per-object depth). To conduct a fair comparison with the existing benchmark in \cite{zhu2019learning}, we follow its experimental setting to use the groundtruth 2D boxes (to filter out the impact of 2D object detection) in all datasets. We conduct data association with the Kalman-filter based tracker in \cite{bewley2016simple}. The comparison on the Waymo Open Dataset, KITTI Detection Dataset, and KITTI MOT Dataset are shown in Table~\ref{tab:compare-waymo}, Table~\ref{tab:compare-kitti-detection}, and Table~\ref{tab:compare-kitti-mot}.

We first demonstrate the effectiveness of the PR-Fusion on both the Waymo Open Dataset and KITTI Detection dataset: with access of two different representations, our methods with PR-Fusion (Ours(PR)) significantly enhances the performance compared to the two baseline models individually, which suggests that the two types of representations are indeed complementary with each other, and the fusion of them yields the best performance.
Regarding the T-Fusion with ego-motion compensation and the PRT-Fusion on the Waymo Open Dataset and the KITTI MOT dataset, when compared to the baseline method PatchNet~\cite{ma2020rethinking}, our proposed method (Ours(T)) achieved significant better performance. Finally, when leverage both the PR-Fusion and T-Fusion together as PRT-Fusion, the performances are further improved (see Ours (PRT)). In summary, our proposed fusion methods show significant improvements over the baseline models and outperform the state-of-the-art methods on both the Waymo Open Dataset and KITTI dataset.

\begin{table}[t]
\centering
\caption{The performance comparison of different fusion strategies on the Waymo Open Dataset for vehicles.}
\scalebox{0.9}{
\begin{tabular}{c|cccc}
\hline
Method & $\absrel$ & $\sqrel$ & $\rmse$ & $\rmselog$ \\  
\hline
\multicolumn{5}{c}{Ablation Study with Predicted Association} \\
\hline
$\text{RGB}_{\text{t}}$  &7.25\% &0.6240 &4.7506 &0.1146\\
$\text{T-Fusion:RGB}_{\text{t}-1\rightarrow \text{t}}$ &7.58\% &0.6773 &4.8902 &0.1172\\
$\text{T-Fusion:RGB}_{\text{t}-3\rightarrow \text{t}}$  &7.60\% &0.6748 &4.9320 &0.1179\\
\hline
$\text{PL}_{\text{t}}$  &8.77\% &0.6051 &5.5485 &0.1283 \\
$\text{T-Fusion:PL}_{\text{t}-1\rightarrow \text{t}}$  &6.96\% &0.4214 &4.5941 &0.1001\\
$\text{T-Fusion:PL}_{\text{t}-3\rightarrow \text{t}}$ &7.16\% &0.4389 &4.6712 &0.1022\\
\hline
$\text{PRT: RGB}_{\text{t}} + \text{PL}_{ \text{t}}$   &5.74\% &0.3188 &3.8788 &0.0863\\
$\text{PRT: RGB}_{\text{t}} + \text{PL}_{\text{t}-1\rightarrow \text{t}}$   &\textbf{5.47\%} &\textbf{0.2858} &\textbf{3.7282} &\textbf{0.0802}\\
$\text{PRT: RGB}_{\text{t}} + \text{PL}_{\text{t}-3\rightarrow \text{t}}$  &5.52\% &0.2932 &3.7661 &0.0807\\
\hline
\multicolumn{5}{c}{Ablation Study with Groundtruth Association} \\
\hline
$\text{PRT: RGB}_{\text{t}} + \text{PL}_{ \text{t}}$ &5.74\% &0.3188 &3.8788 &0.0863 \\
$\text{PRT: RGB}_{\text{t}} + \text{PL}_{\text{t}-1\rightarrow \text{t}}$  &5.34\% &0.2713 &3.6278 &0.0791\\
$\text{PRT: RGB}_{\text{t}} + \text{PL}_{\text{t}-3\rightarrow \text{t}}$  &\textbf{5.29\%} &\textbf{0.2668} &\textbf{3.5821} &\textbf{0.0783}\\
\hline
\end{tabular}
}
\vspace{-8pt}
\label{tab:fusion-ablation}
\end{table}

\subsection{Ablation Study for Per-object Depth Estimation}
\label{sec:exp:ablation}

To better demonstrate and understand the effectiveness of each module of our proposed method, we conducted thorough ablation studies with different fusion strategies and different 2D boxes and tracking qualities.

{
For the first three ablation studies, we report performance with predicted association to better understand how our method works in practice, and in the fourth study we conduct headroom analysis to understand how would our method work given perfect association.

1. \textbf{Does tracklet fusion work for RGB features?} We simply fused the features extracted from RGB images from a tracklet (similar with our proposed T-Fusion, but without ego-motion compensation), and results are shown in the first group of Table~\ref{tab:fusion-ablation} and only marginal improvements are observed. One possible explanation is that the 3D information encoded in RGB images at different timestamps are at different coordinate system if the self-driving car is moving. It is non-trivial to decompose the camera ego motion and the object motion, which makes it hard to learn motion and temporal consistency from simply fusing the image features.

2. \textbf{How does tracklet fusion improve pseudo-LiDAR-based depth estimation?} In the second group of Table \ref{tab:fusion-ablation} we show the performance of pseudo-LiDAR-based depth estimation with our proposed ego-motion compensation based T-Fusion. It is clearly that the depth estimation performance significantly improved with the T-Fusion even with just information from one more frame. Adding more frames is helpful, but the improvement is marginal.

3. \textbf{Does PRT fusion help?} With the improvements of PR-Fusion and T-Fusion, it is natural to ask the question about if the combination of them is helpful. The third group of Table~\ref{tab:fusion-ablation} shows the performance of the PRT-Fusion with different number of frames. It is clear that the combination of both (PR-Fusion and T-Fusion) outperforms each one individually. However, due to noise introduced by data association, a longer tracklet does not necessarily yield better performance.

4. \textbf{How is the performance affected by association noise?} We further study the impact of the association quality towards our proposed method. The fourth group shows the results with the groundtruth association. As shown in Table.~\ref{tab:fusion-ablation}, we can observe with perfect association, the improvement is consistent with longer tracklet. This indicates that to fully leverage the capability of the proposed fusion method, improving data association quality is a promising direction to work on.
}

\subsection{Improving Monocular 3D Detection and Tracking with Enhanced Per-object Depth}
\label{sec:exp:app}

In this subsection, we apply our per-object depth estimation to show that it can further help improve the state-of-the-art monocular image based 3D detector CenterNet\cite{zhou2020tracking} and the AB3D tracker \cite{weng20203d} on the Waymo Open Dataset.

\textbf{Quantitative results.} For 3D object detection, we trained a CenterNet for monocular 3D detection, and replace only the depth of the detection results while the other outputs such as box, rotation, etc. are remained. For tracking, we follow the tracking-by-detection scheme and perform a AB3D tracker \cite{weng20203d} on the detection boxes with the enhanced per-object depth. The results of detection and tracking are shown in table~\ref{tab:detection} and table~\ref{tab:tracking}. As expected, we can observe that by simply enhancing the depth, significant improvements can be obtained for both tasks. It is worth noting that the performance can be further improved by tuning the model specifically for detection and tracking, but it is out of the scope of this paper since we only focus on demonstrating the improvements purely from depth. 

\begin{table}[t!]
\caption{The results of monocular 3D detection on the Waymo Open Dataset for vehicles with different IOU thresholds ($0.5$-$0.7$).}
\begin{center}
\resizebox{0.95\linewidth}{!}{
\begin{tabular}{c|ccc|ccc}
\hline
\multirow{2}{*}{ } & \multicolumn{3}{c|}{AP$_\text{3D}$} & \multicolumn{3}{c}{AP$_\text{BEV}$}\\ 
& 0.5 & 0.6 & 0.7 & 0.5 & 0.6 & 0.7 \\ 
\hline
CenterNet~\cite{zhou2019objects}  &24.32 &14.36 &6.06 &28.20 &18.71 &11.52\\
\multirow{2}{*}{+Enhanced Depth}&27.06 &16.89 &7.99 &29.53 &24.16 &14.33\\
&\textbf{\improves{+2.74}} &\textbf{\improves{+2.53}} &\textbf{\improves{+1.93}} &\textbf{\improves{+1.10}} &\textbf{\improves{+5.45}} &\textbf{\improves{+2.81}}\\
\hline
\end{tabular}
}
\end{center}
\label{tab:detection}
\end{table}

\begin{table}[t]
\begin{center}
\caption{Results on the Waymo Open Dataset for vehicles in~\cite{weng20203d}.}
\resizebox{0.95\linewidth}{!}{
\begin{tabular}{c|c|c|c|c|c}
\hline
Method       & SMOTA $\higherbetter$ &AMOTA $\higherbetter$ & AMOTP $\higherbetter$ & MOTA $\higherbetter$ &IDS $\lowerbetter$\\
\hline
CenterNet~\cite{zhou2019objects} + AB3D\cite{weng20203d} &50.19 &14.06 &27.00 &39.37 &228\\

\multirow{2}{*}{+Enhanced Depth}&52.49 &15.24 &28.50 &41.24 &130\\
&\textbf{\improves{+2.30}} &\textbf{\improves{+1.18}} &\textbf{\improves{+1.50}} &\textbf{\improves{+1.87}} &\textbf{\improves{-98}}\\
\hline
\end{tabular}
}
\end{center}
\vspace{-10pt}
\label{tab:tracking}
\end{table}

\begin{figure}
\begin{center}
\includegraphics[width=0.6\linewidth]{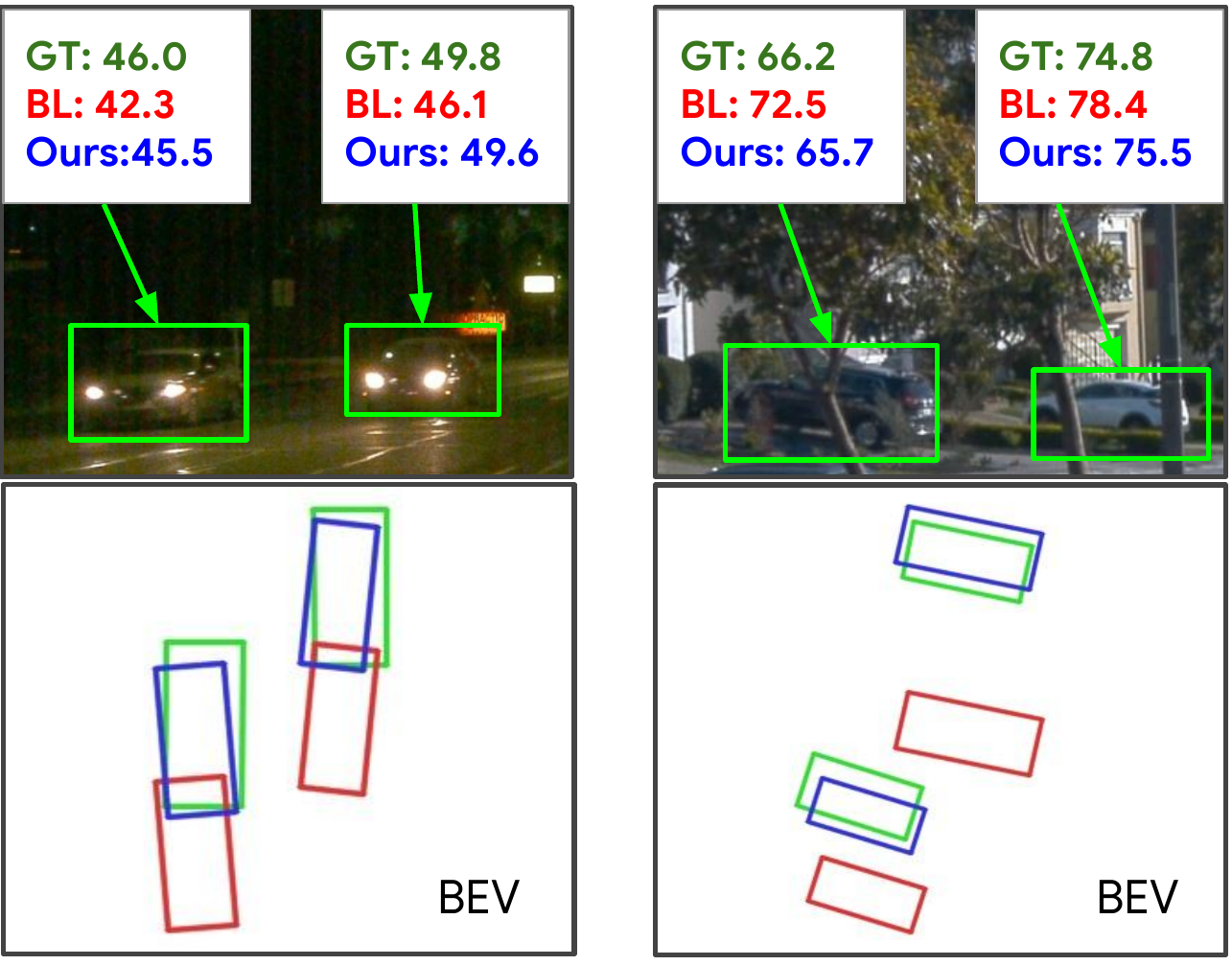}
\end{center}
   \caption{Qualitative examples of per-object depth estimation and monocular 3D object detection. The green, red, and blue bounding boxes corresponding to ground truth (GT), baseline depth estimation and detection (BL), and the one with the enhanced per-object depth from our proposed PRT-Fusion. Significant better depth estimation and its further improvements on detection can be observed.}
\label{fig:detection-visualization}
\end{figure}

\begin{figure}
\begin{center}
\includegraphics[width=0.8\linewidth]{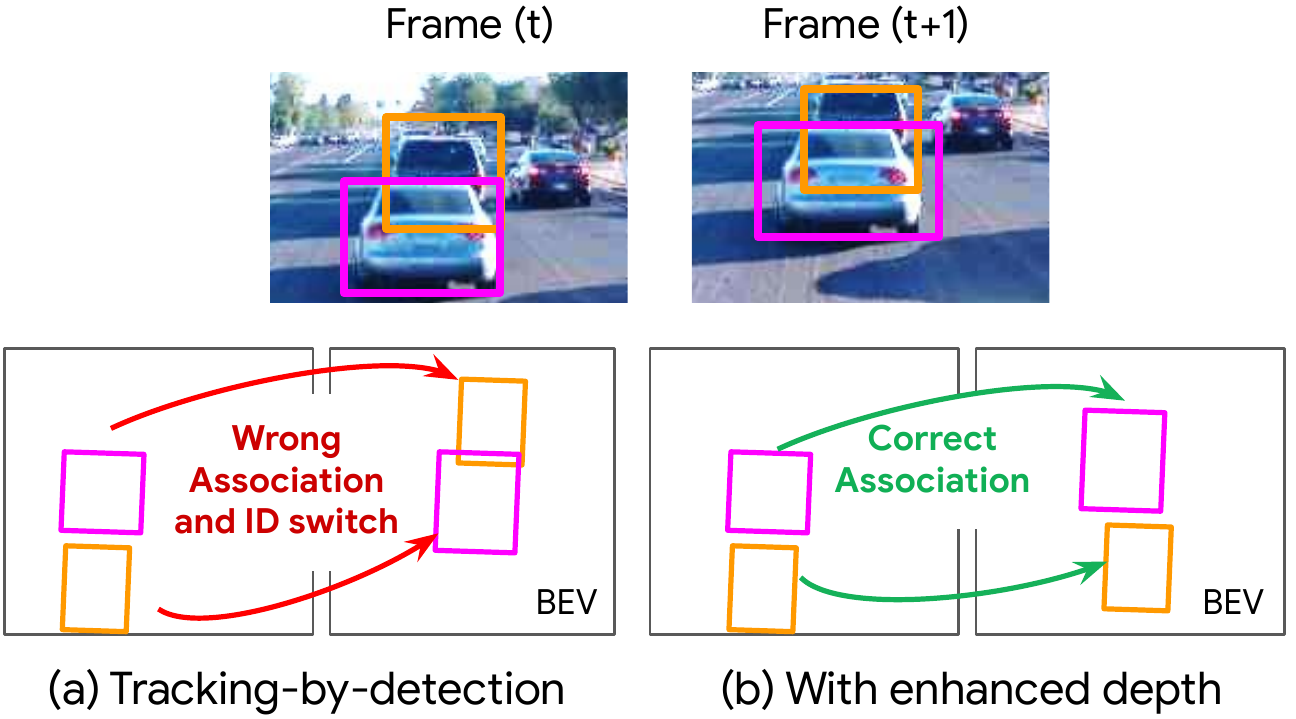}
\end{center}
\caption{Qualitative examples of monocular 3D tracking results. Due to inaccurate depth estimation show in (a), the 3D tracker wrongly associates detection across frames which leads to ID switches. With enhanced depth predicted by our proposed fusion model in (b), the tracker associates detections correctly.}
\label{fig:tracking-visualization}
\end{figure}

\textbf{Qualitative results.} We further visualize the prediction results of the baseline detection model and our proposed PRT-Fusion to illustrate how the enhanced per-object depth improve the monocular 3D detection and tracking. As shown in Fig.~\ref{fig:detection-visualization}, the first row shows the improvements of per-object depth brought by our method, and the second row illustrated the bird-eye-view of the 3D detection results of both the baseline detector, and the one with our improved depth (only the depth is replaced). Fig.~\ref{fig:tracking-visualization} shows the tracking model makes fewer ID Switch errors with the depth predicted by our model. Clear improvements can be observed, and the improved depth is the key factor to lead to the significant improvements for monocular 3D detection and tracking.   

\section{Conclusion}
We demonstrated that per-object depth estimation is the performance bottleneck of the monocular image based 3D perception tasks including detection and tracking. A multi-level fusion framework was proposed to fuse features from different representations across multiple frames. We first obtained the state-of-the-arts performance on per-object depth estimation, and then showed that by simply replacing the depth, significant improvements can be observed in the tasks above. This not only demonstrated our findings and the effectiveness of the proposed method, but also indicating that improving per-object depth is a promising direction to enhance detection and tracking. Future works can include end-to-end training of the proposed method.

\section{Acknowledgement}

We would like to thank Jiyang Gao for the helpful discussions about this work.

{\small
\bibliographystyle{IEEEtran}
\bibliography{icra}

\begin{thebibliography}{10}
\providecommand{\url}[1]{#1}
\csname url@samestyle\endcsname
\providecommand{\newblock}{\relax}
\providecommand{\bibinfo}[2]{#2}
\providecommand{\BIBentrySTDinterwordspacing}{\spaceskip=0pt\relax}
\providecommand{\BIBentryALTinterwordstretchfactor}{4}
\providecommand{\BIBentryALTinterwordspacing}{\spaceskip=\fontdimen2\font plus
\BIBentryALTinterwordstretchfactor\fontdimen3\font minus
  \fontdimen4\font\relax}
\providecommand{\BIBforeignlanguage}[2]{{%
\expandafter\ifx\csname l@#1\endcsname\relax
\typeout{** WARNING: IEEEtran.bst: No hyphenation pattern has been}%
\typeout{** loaded for the language `#1'. Using the pattern for}%
\typeout{** the default language instead.}%
\else
\language=\csname l@#1\endcsname
\fi
#2}}
\providecommand{\BIBdecl}{\relax}
\BIBdecl

\bibitem{lang2019pointpillars}
A.~H. Lang, S.~Vora, H.~Caesar, L.~Zhou, J.~Yang, and O.~Beijbom,
  ``Pointpillars: Fast encoders for object detection from point clouds,'' in
  \emph{Proceedings of the IEEE Conference on Computer Vision and Pattern
  Recognition}, 2019, pp. 12\,697--12\,705.

\bibitem{liu2019flownet3d}
X.~Liu, C.~R. Qi, and L.~J. Guibas, ``Flownet3d: Learning scene flow in 3d
  point clouds,'' in \emph{Proceedings of the IEEE Conference on Computer
  Vision and Pattern Recognition}, 2019, pp. 529--537.

\bibitem{behley2019semantickitti}
J.~Behley, M.~Garbade, A.~Milioto, J.~Quenzel, S.~Behnke, C.~Stachniss, and
  J.~Gall, ``Semantickitti: A dataset for semantic scene understanding of lidar
  sequences,'' in \emph{Proceedings of the IEEE International Conference on
  Computer Vision}, 2019, pp. 9297--9307.

\bibitem{yang2020radarnet}
B.~Yang, R.~Guo, M.~Liang, S.~Casas, and R.~Urtasun, ``Radarnet: Exploiting
  radar for robust perception of dynamic objects,'' \emph{arXiv preprint
  arXiv:2007.14366}, 2020.

\bibitem{chen2020monopair}
Y.~Chen, L.~Tai, K.~Sun, and M.~Li, ``Monopair: Monocular 3d object detection
  using pairwise spatial relationships,'' in \emph{Proceedings of the IEEE/CVF
  Conference on Computer Vision and Pattern Recognition}, 2020, pp.
  12\,093--12\,102.

\bibitem{alhashim2018high}
I.~Alhashim and P.~Wonka, ``High quality monocular depth estimation via
  transfer learning,'' \emph{arXiv preprint arXiv:1812.11941}, 2018.

\bibitem{wang2019pseudo}
Y.~Wang, W.-L. Chao, D.~Garg, B.~Hariharan, M.~Campbell, and K.~Q. Weinberger,
  ``Pseudo-lidar from visual depth estimation: Bridging the gap in 3d object
  detection for autonomous driving,'' in \emph{Proceedings of the IEEE
  Conference on Computer Vision and Pattern Recognition}, 2019, pp. 8445--8453.

\bibitem{ma2020rethinking}
X.~Ma, S.~Liu, Z.~Xia, H.~Zhang, X.~Zeng, and W.~Ouyang, ``Rethinking
  pseudo-lidar representation,'' \emph{arXiv preprint arXiv:2008.04582}, 2020.

\bibitem{zhou2019objects}
X.~Zhou, D.~Wang, and P.~Kr{\"a}henb{\"u}hl, ``Objects as points,'' \emph{arXiv
  preprint arXiv:1904.07850}, 2019.

\bibitem{weng20203d}
X.~Weng, J.~Wang, D.~Held, and K.~Kitani, ``3d multi-object tracking: A
  baseline and new evaluation metrics,'' \emph{arXiv preprint
  arXiv:1907.03961}, 2020.

\bibitem{hu2019joint}
H.-N. Hu, Q.-Z. Cai, D.~Wang, J.~Lin, M.~Sun, P.~Krahenbuhl, T.~Darrell, and
  F.~Yu, ``Joint monocular 3d vehicle detection and tracking,'' in
  \emph{Proceedings of the IEEE international conference on computer vision},
  2019, pp. 5390--5399.

\bibitem{wang2020centernet3d}
G.~Wang, B.~Tian, Y.~Ai, T.~Xu, L.~Chen, and D.~Cao, ``Centernet3d: An anchor
  free object detector for autonomous driving,'' \emph{arXiv preprint
  arXiv:2007.07214}, 2020.

\bibitem{RTM3D}
P.~Li, H.~Zhao, P.~Liu, and F.~Cao, ``Rtm3d: Real-time monocular 3d detection
  from object keypoints for autonomous driving,'' in \emph{ECCV}, 2020.

\bibitem{zhou2020tracking}
X.~Zhou, V.~Koltun, and P.~Kr{\"a}henb{\"u}hl, ``Tracking objects as points,''
  \emph{ECCV}, 2020.

\bibitem{caesar2020nuscenes}
H.~Caesar, V.~Bankiti, A.~H. Lang, S.~Vora, V.~E. Liong, Q.~Xu, A.~Krishnan,
  Y.~Pan, G.~Baldan, and O.~Beijbom, ``nuscenes: A multimodal dataset for
  autonomous driving,'' in \emph{Proceedings of the IEEE/CVF Conference on
  Computer Vision and Pattern Recognition}, 2020, pp. 11\,621--11\,631.

\bibitem{liu2020smoke}
Z.~Liu, Z.~Wu, and R.~T{\'o}th, ``Smoke: Single-stage monocular 3d object
  detection via keypoint estimation,'' in \emph{Proceedings of the IEEE/CVF
  Conference on Computer Vision and Pattern Recognition Workshops}, 2020, pp.
  996--997.

\bibitem{vianney2019refinedmpl}
J.~M.~U. Vianney, S.~Aich, and B.~Liu, ``Refinedmpl: Refined monocular
  pseudolidar for 3d object detection in autonomous driving,'' \emph{arXiv
  preprint arXiv:1911.09712}, 2019.

\bibitem{godard2017unsupervised}
C.~Godard, O.~Mac~Aodha, and G.~J. Brostow, ``Unsupervised monocular depth
  estimation with left-right consistency,'' in \emph{Proceedings of the IEEE
  Conference on Computer Vision and Pattern Recognition}, 2017, pp. 270--279.

\bibitem{godard2019digging}
C.~Godard, O.~Mac~Aodha, M.~Firman, and G.~J. Brostow, ``Digging into
  self-supervised monocular depth estimation,'' in \emph{Proceedings of the
  IEEE international conference on computer vision}, 2019, pp. 3828--3838.

\bibitem{simonyan2014two}
K.~Simonyan and A.~Zisserman, ``Two-stream convolutional networks for action
  recognition in videos,'' in \emph{Advances in neural information processing
  systems}, 2014, pp. 568--576.

\bibitem{qi2018frustum}
C.~R. Qi, W.~Liu, C.~Wu, H.~Su, and L.~J. Guibas, ``Frustum pointnets for 3d
  object detection from rgb-d data,'' in \emph{Proceedings of the IEEE
  conference on computer vision and pattern recognition}, 2018, pp. 918--927.

\bibitem{beery2020context}
S.~Beery, G.~Wu, V.~Rathod, R.~Votel, and J.~Huang, ``Context r-cnn: Long term
  temporal context for per-camera object detection,'' in \emph{Proceedings of
  the IEEE/CVF Conference on Computer Vision and Pattern Recognition}, 2020,
  pp. 13\,075--13\,085.

\bibitem{liu2019looking}
M.~Liu, M.~Zhu, M.~White, Y.~Li, and D.~Kalenichenko, ``Looking fast and slow:
  Memory-guided mobile video object detection,'' \emph{arXiv preprint
  arXiv:1903.10172}, 2019.

\bibitem{zhu2017flow}
X.~Zhu, Y.~Wang, J.~Dai, L.~Yuan, and Y.~Wei, ``Flow-guided feature aggregation
  for video object detection,'' in \emph{Proceedings of the IEEE International
  Conference on Computer Vision}, 2017, pp. 408--417.

\bibitem{sun2020scalability}
P.~Sun, H.~Kretzschmar, X.~Dotiwalla, A.~Chouard, V.~Patnaik, P.~Tsui, J.~Guo,
  Y.~Zhou, Y.~Chai, B.~Caine \emph{et~al.}, ``Scalability in perception for
  autonomous driving: Waymo open dataset,'' in \emph{Proceedings of the
  IEEE/CVF Conference on Computer Vision and Pattern Recognition}, 2020, pp.
  2446--2454.

\bibitem{geiger2013vision}
A.~Geiger, P.~Lenz, C.~Stiller, and R.~Urtasun, ``Vision meets robotics: The
  kitti dataset,'' \emph{The International Journal of Robotics Research},
  vol.~32, no.~11, pp. 1231--1237, 2013.

\bibitem{zhu2019learning}
J.~Zhu and Y.~Fang, ``Learning object-specific distance from a monocular
  image,'' in \emph{Proceedings of the IEEE International Conference on
  Computer Vision}, 2019, pp. 3839--3848.

\bibitem{brazil2019m3d}
G.~Brazil and X.~Liu, ``M3d-rpn: Monocular 3d region proposal network for
  object detection,'' in \emph{Proceedings of the IEEE International Conference
  on Computer Vision}, 2019, pp. 9287--9296.

\bibitem{simonelli2019disentangling}
A.~Simonelli, S.~R. Bulo, L.~Porzi, M.~L{\'o}pez-Antequera, and
  P.~Kontschieder, ``Disentangling monocular 3d object detection,'' in
  \emph{Proceedings of the IEEE International Conference on Computer Vision},
  2019, pp. 1991--1999.

\bibitem{jorgensen2019monocular}
E.~J{\"o}rgensen, C.~Zach, and F.~Kahl, ``Monocular 3d object detection and box
  fitting trained end-to-end using intersection-over-union loss,'' \emph{arXiv
  preprint arXiv:1906.08070}, 2019.

\bibitem{weng2019monocular}
X.~Weng and K.~Kitani, ``Monocular 3d object detection with pseudo-lidar point
  cloud,'' in \emph{Proceedings of the IEEE International Conference on
  Computer Vision Workshops}, 2019, pp. 0--0.

\bibitem{ding2020learning}
M.~Ding, Y.~Huo, H.~Yi, Z.~Wang, J.~Shi, Z.~Lu, and P.~Luo, ``Learning
  depth-guided convolutions for monocular 3d object detection,'' in
  \emph{Proceedings of the IEEE/CVF Conference on Computer Vision and Pattern
  Recognition Workshops}, 2020, pp. 1000--1001.

\bibitem{ma2019accurate}
X.~Ma, Z.~Wang, H.~Li, P.~Zhang, W.~Ouyang, and X.~Fan, ``Accurate monocular 3d
  object detection via color-embedded 3d reconstruction for autonomous
  driving,'' in \emph{Proceedings of the IEEE International Conference on
  Computer Vision}, 2019, pp. 6851--6860.

\bibitem{bewley2016simple}
A.~Bewley, Z.~Ge, L.~Ott, F.~Ramos, and B.~Upcroft, ``Simple online and
  realtime tracking,'' in \emph{2016 IEEE International Conference on Image
  Processing (ICIP)}.\hskip 1em plus 0.5em minus 0.4em\relax IEEE, 2016, pp.
  3464--3468.

\bibitem{weng2020gnn3dmot}
X.~Weng, Y.~Wang, Y.~Man, and K.~M. Kitani, ``Gnn3dmot: Graph neural network
  for 3d multi-object tracking with 2d-3d multi-feature learning,'' in
  \emph{Proceedings of the IEEE/CVF Conference on Computer Vision and Pattern
  Recognition}, 2020, pp. 6499--6508.

\bibitem{eigen2014depth}
D.~Eigen, C.~Puhrsch, and R.~Fergus, ``Depth map prediction from a single image
  using a multi-scale deep network,'' in \emph{Advances in neural information
  processing systems}, 2014, pp. 2366--2374.

\bibitem{guizilini2020semantically}
V.~Guizilini, R.~Hou, J.~Li, R.~Ambrus, and A.~Gaidon, ``Semantically-guided
  representation learning for self-supervised monocular depth,'' \emph{arXiv
  preprint arXiv:2002.12319}, 2020.

\bibitem{wang2020task}
X.~Wang, W.~Yin, T.~Kong, Y.~Jiang, L.~Li, and C.~Shen, ``Task-aware monocular
  depth estimation for 3d object detection.'' in \emph{AAAI}, 2020, pp.
  12\,257--12\,264.

\bibitem{poggi2020uncertainty}
M.~Poggi, F.~Aleotti, F.~Tosi, and S.~Mattoccia, ``On the uncertainty of
  self-supervised monocular depth estimation,'' in \emph{Proceedings of the
  IEEE/CVF Conference on Computer Vision and Pattern Recognition}, 2020, pp.
  3227--3237.

\bibitem{fu2018deep}
H.~Fu, M.~Gong, C.~Wang, K.~Batmanghelich, and D.~Tao, ``Deep ordinal
  regression network for monocular depth estimation,'' in \emph{Proceedings of
  the IEEE Conference on Computer Vision and Pattern Recognition}, 2018, pp.
  2002--2011.

\bibitem{vora2020pointpainting}
S.~Vora, A.~H. Lang, B.~Helou, and O.~Beijbom, ``Pointpainting: Sequential
  fusion for 3d object detection,'' in \emph{Proceedings of the IEEE/CVF
  Conference on Computer Vision and Pattern Recognition}, 2020, pp. 4604--4612.

\bibitem{liang2019multi}
M.~Liang, B.~Yang, Y.~Chen, R.~Hu, and R.~Urtasun, ``Multi-task multi-sensor
  fusion for 3d object detection,'' in \emph{Proceedings of the IEEE Conference
  on Computer Vision and Pattern Recognition}, 2019, pp. 7345--7353.

\bibitem{liang2018deep}
M.~Liang, B.~Yang, S.~Wang, and R.~Urtasun, ``Deep continuous fusion for
  multi-sensor 3d object detection,'' in \emph{Proceedings of the European
  Conference on Computer Vision (ECCV)}, 2018, pp. 641--656.

\bibitem{fadadu2020multi}
S.~Fadadu, S.~Pandey, D.~Hegde, Y.~Shi, F.-C. Chou, N.~Djuric, and
  C.~Vallespi-Gonzalez, ``Multi-view fusion of sensor data for improved
  perception and prediction in autonomous driving,'' \emph{arXiv preprint
  arXiv:2008.11901}, 2020.

\bibitem{xiao2020audiovisual}
F.~Xiao, Y.~J. Lee, K.~Grauman, J.~Malik, and C.~Feichtenhofer, ``Audiovisual
  slowfast networks for video recognition,'' \emph{arXiv preprint
  arXiv:2001.08740}, 2020.

\bibitem{hazirbas2016fusenet}
C.~Hazirbas, L.~Ma, C.~Domokos, and D.~Cremers, ``Fusenet: Incorporating depth
  into semantic segmentation via fusion-based cnn architecture,'' in
  \emph{Asian conference on computer vision}.\hskip 1em plus 0.5em minus
  0.4em\relax Springer, 2016, pp. 213--228.

\bibitem{feichtenhofer2016convolutional}
C.~Feichtenhofer, A.~Pinz, and A.~Zisserman, ``Convolutional two-stream network
  fusion for video action recognition,'' in \emph{Proceedings of the IEEE
  conference on computer vision and pattern recognition}, 2016, pp. 1933--1941.

\bibitem{dosovitskiy2015flownet}
A.~Dosovitskiy, P.~Fischer, E.~Ilg, P.~Hausser, C.~Hazirbas, V.~Golkov, P.~Van
  Der~Smagt, D.~Cremers, and T.~Brox, ``Flownet: Learning optical flow with
  convolutional networks,'' in \emph{Proceedings of the IEEE international
  conference on computer vision}, 2015, pp. 2758--2766.

\bibitem{he2016deep}
K.~He, X.~Zhang, S.~Ren, and J.~Sun, ``Deep residual learning for image
  recognition,'' in \emph{Proceedings of the IEEE conference on computer vision
  and pattern recognition}, 2016, pp. 770--778.

\bibitem{zhang2020fair}
Y.~Zhang, C.~Wang, X.~Wang, W.~Zeng, and W.~Liu, ``Fairmot: On the fairness of
  detection and re-identification in multiple object tracking,'' \emph{arXiv
  preprint arXiv:2004.01888}, 2020.

\bibitem{hung2020soda}
W.-C. Hung, H.~Kretzschmar, T.-Y. Lin, Y.~Chai, R.~Yu, M.-H. Yang, and
  D.~Anguelov, ``Soda: Multi-object tracking with soft data association,''
  \emph{arXiv preprint arXiv:2008.07725}, 2020.

\bibitem{wang2019towards}
Z.~Wang, L.~Zheng, Y.~Liu, Y.~Li, and S.~Wang, ``Towards real-time multi-object
  tracking,'' \emph{arXiv preprint arXiv:1909.12605}, 2019.

\bibitem{gokcce2015vision}
F.~G{\"o}k{\c{c}}e, G.~{\"U}{\c{c}}oluk, E.~{\c{S}}ahin, and S.~Kalkan,
  ``Vision-based detection and distance estimation of micro unmanned aerial
  vehicles,'' \emph{Sensors}, vol.~15, no.~9, pp. 23\,805--23\,846, 2015.

\bibitem{haseeb2018disnet}
M.~A. Haseeb, J.~Guan, D.~Risti{\'c}-Durrant, and A.~Gr{\"a}ser, ``Disnet: A
  novel method for distance estimation from monocular camera,'' \emph{10th
  Planning, Perception and Navigation for Intelligent Vehicles (PPNIV18),
  IROS}, 2018.

\bibitem{tuohy2010distance}
S.~Tuohy, D.~O'Cualain, E.~Jones, and M.~Glavin, ``Distance determination for
  an automobile environment using inverse perspective mapping in opencv,''
  2010.

\end{thebibliography}
}

\end{document}